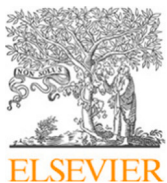
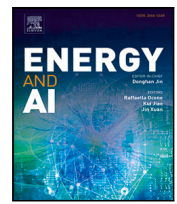

# Physics-Guided Memory Network for building energy modeling

Muhammad Umair Danish [a], Kashif Ali [b], Kamran Siddiqui [b], Katarina Grolinger [a],*

[a] *Electrical and Computer Engineering, Western University, 1151 Richmond St, London, N6A 3K7, Ontario, Canada*
[b] *Mechanical and Materials Engineering, Western University, 1151 Richmond St, London, N6A 3K7, Ontario, Canada*

## HIGHLIGHTS

- Designed Physics-Guided Memory Network (PgMN) for building energy prediction.
- PgMN combines the advantages of both physics-based and machine-learning models.
- PgMN adapts to new and modified buildings, utilizing available historical data.
- Designed Parallel Projection Layers, Memory Unit, and Memory Experience Module.
- Provided theoretical proofs and experimental validation of PgMN performance.

## GRAPHICAL ABSTRACT

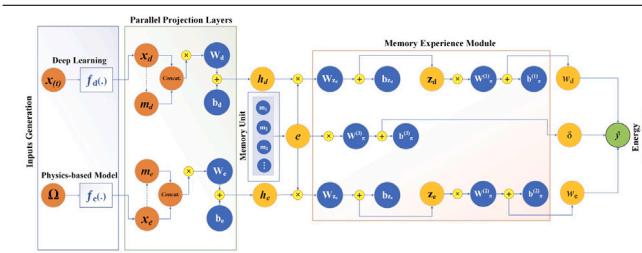

The Physics-Guided Memory Network architecture comprises four primary components: Input Generations, Parallel Projection Layers, Memory Unit, and Memory Experience Module. The Parallel Projection Layers transform input features into learned representations, while the Memory Unit captures persistent biases, and the Memory Experience Module optimally combines inputs to produce accurate energy forecasts. The Memory Unit is a vector across training instances for capturing forecasting biases in a learnable form.

## ARTICLE INFO

## ABSTRACT

Accurate energy consumption forecasting is essential for efficient resource management and sustainability in the building sector. Deep learning models are highly successful but struggle with limited historical data and become unusable when historical data are unavailable, such as in newly constructed buildings. On the other hand, physics-based models, such as EnergyPlus, simulate energy consumption without relying on historical data but require extensive building parameter specifications and considerable time to model a building. This paper introduces a Physics-Guided Memory Network (PgMN), a neural network that integrates predictions from deep learning and physics-based models to address their limitations. PgMN comprises a Parallel Projection Layers to process incomplete inputs, a Memory Unit to account for persistent biases, and a Memory Experience Module to optimally extend forecasts beyond their input range and produce output. Theoretical evaluation shows that components of PgMN are mathematically valid for performing their respective tasks. The PgMN was evaluated on short-term energy forecasting at an hourly resolution, critical for operational decision-making in smart grid and smart building systems. Experimental validation shows accuracy and applicability of PgMN in diverse scenarios such as newly constructed buildings, missing data, sparse historical data, and dynamic infrastructure changes. This paper provides a promising solution for energy consumption forecasting in dynamic building environments, enhancing model applicability in scenarios where historical data are limited or unavailable or when physics-based models are inadequate.

☆ This work was supported in part by the Climate Action and Awareness Fund [EDF-CA-2021i018, Environnement Canada, K. Siddiqui and K. Grolinger] and in part by the Canada Research Chairs Program [CRC-2022-00078, K. Grolinger].
* Corresponding author.
 *E-mail address:* kgroling@uwo.ca (K. Grolinger).

https://doi.org/10.1016/j.egyai.2025.100538
Received 3 February 2025; Received in revised form 3 May 2025; Accepted 13 June 2025
Available online 29 June 2025

## 1. Introduction

Deep learning (DL) models in the energy sector have proven transformative, optimizing resource management, enabling data-driven decision-making, and advancing sustainability goals [1]. These DL models have shown the potential to improve prediction accuracy by up to 25% while simultaneously reducing operational costs by as much as 30% [2]. Alongside DL models, physics-based models, such as Building Energy Models (BEMs), remain essential to building energy modeling by providing detailed insights into the energy performance of buildings. These simulation programs become more important than data-driven models when historical data are incomplete or unavailable and play a crucial role in early-stage building design and optimization.

Despite advancements in energy consumption forecasting, accurately predicting energy remains a challenge due to several factors, such as envelope or infrastructure changes due to building renovations, equipment changes, variations in meteorological conditions, and the behavior of building occupants [3]. DL models have been successful; however, since they are inherently data-driven and rely on historical data for training, they cannot be deployed for newly constructed buildings without data. Similarly, DL models are infeasible and ineffective when historical data for training are incomplete. In contrast, BEMs are principle-driven, do not require historical data, and can be deployed for newly constructed buildings [4,5]. The changes in building infrastructure, such as HVAC (Heating, Ventilation, and Air Conditioning) modifications, can reduce the accuracy of DL models because they rely on historical training data and do not have any physical information related to the building. In comparison, BEMs leverage detailed physical characteristics of buildings, such as building envelopes, lighting, appliances, and HVAC systems. However, modeling this information in BEMs requires substantial time and effort. Moreover, one of the most influential factors impacting system performance is consumer energy usage behavior, such as routines, where BEMs face challenges as occupancy schedules must be known. These challenges require designing a system that combines the benefits of both DL and BEMs to enhance applicability in diverse situations.

Simulated energy consumption data produced by BEMs presents several advantages that can complement and enhance DL models. For instance, BEM can generate simulated data to train DL models for newly constructed buildings where historical data is unavailable. Additionally, in cases of sensor failures, communication errors, or maintenance downtime, where historical data are incomplete or insufficient, BEM simulations can provide valuable data to support the learning process of DL models. Moreover, when there are changes in building infrastructure, such as updates to HVAC systems, these alterations can affect energy consumption, and DL models lack prior knowledge of such changes. However, BEMs can be updated to reflect the new configurations and aid DL models in adjusting their forecasts based on new physical changes in the building. Moreover, BEMs such as EnergyPlus (EP), can incorporate human behavioral patterns, including seasonal variations and changes in daily routines (e.g., altered energy usage during winter), to enhance the accuracy of energy performance simulations.

Despite the considerable benefits of combining predictions from BEMs and DL models, their integration has not been explored to enhance building energy consumption modeling performance. However, parallels can be drawn from successful applications in other domains where physics-based models and DL were combined as Physics-Guided Neural Networks (PgNN) to achieve substantial advancements. For example, Singh et al. [6] utilized a physics-guided hybrid method to model dynamic systems such as inverted pendulums and tumor growth, stressing the potential of integrating Physics-Based Model (PBM) outputs into DL models. Similarly, Jia et al. [7] incorporated density-depth relationships, a physics-based principle, into the Long Short-Term Memory (LSTM) model to improve lake temperature predictions in the absence of complete data. In computer vision, PBMs such as Multi-joint Dynamics with Contact (MuJoCo) have been leveraged to generate low-fidelity images, which are then used to train Convolutional Neural Networks (CNNs) for object detection [8]. These instances stress that integrating BEM outputs, such as those from EnergyPlus, with DL models could address limitations in energy consumption forecasting, enabling more accurate predictions that adapt to changes in building infrastructure, usage patterns, and environmental conditions.

DL models and BEMs have independently shown their utility in energy consumption forecasting and modeling. However, when applied individually, these methods have inherent limitations [9]. The DL models struggle in several situations, such as newly constructed buildings or infrastructure changes, where historical data is unavailable or irrelevant. On the other hand, BEMs rely on physical principles and detailed building configurations to simulate physically aligned energy consumption, enabling predictions even for new or modified buildings. However, white-box BEMs usually require extensive setup time and do not learn from real-time historical data in the same way DL models do [10]. BEMs do incorporate historical data through a calibration process. For example, calibration may involve adjusting parameters such as material properties, occupancy schedules, and HVAC settings to align model outputs with real-world energy consumption. This disconnect between the learning ability of DL and the physical rigor of BEMs stresses a gap in energy forecasting and modeling. The opportunity lies in integrating BEMs and DL to combine the strengths of both methods and address diverse scenarios, such as infrastructure modifications, missing sensor data, or dynamic consumer behaviors.

To leverage the strengths of both BEMs and DL, this paper introduces the Physics-Guided Memory Network (PgMN), a neural network that seamlessly integrates predictions from DL models and BEMs to address the inherent limitations of both methods. The proposed approach enables the system to adjust to changing infrastructure, account for consumer behavior shifts, and operate in scenarios with limited or no historical data, providing a robust and accurate solution for energy forecasting in diverse building contexts. PgMN consists of Parallel Projection Layers, a Memory Unit, and a Memory Experience Module. The Parallel Projection Layers transform predictions from DL and BEMs into deep feature space for efficient learning and help locate missing data points. The Memory Unit is a learnable parameter for recording past experiences and aiding the model in preventing biases, allowing the model to dynamically adjust predictions by learning from previous forecasting errors. The Memory Experience Module combines deep features from the Parallel Projection Layers and memory units and establishes the best balance between BEMs and DL predictions where the error is minimized. It also enables predictions beyond the margins of both BEMs and DL when necessary to minimize errors. Evaluation across diverse scenarios of short-term energy forecasting at an hourly resolution demonstrates PgMN's capabilities, highlighting the advantages over pure DL or BEMs. The main contributions of this paper are as follows:

1. Design of the Physics-Guided Memory Network, a neural network that integrates predictions from deep learning and simulated data from physics-based models to predict energy consumption under diverse scenarios.
2. Design of the Parallel Projection Layers (PPL) to generate deep features from both BEMs and DL predictions, Memory Unit (MU) to record experience in a learnable manner, and Memory Experience Module, which takes learned representation from PPL and MU to produce the final output. These components ensure the handling of missing data, dynamic error correction, and the ability to combine or extend predictions beyond input margins for enhanced performance.
3. Theoretical evaluation of PgMN through formal analysis and proofs, including Universal Function Approximation, Bias Correction Capability of the Memory Unit, and Unbounded Output Capability, to ensure that the model has satisfactory theoretical rigor for reliable deployment in real-world scenarios.

4. Experimental validation of PgMN across diverse scenarios, including missing data, lack of historical data, and infrastructure changes, highlighting its superior applicability and accuracy compared to standalone DL and BEM approaches.

The remainder of the paper follows the following structure. Section 2 discusses Related Work, Section 3 describes the proposed Physics Guided Memory Network, Section 4 presents theoretical evaluations, and Section 5 provides Results and Evaluations. Finally, Section 6 concludes the paper.

## 2. Related work

This section discusses Physics-Guided Neural Networks and Building Energy Modeling simulation tools.

### 2.1. Physics-Guided Neural Networks (PgNNs)

The integration of physics-based methods and DL has demonstrated success across various domains, including energy systems [11]. This subsection first reviews these applications in other fields, followed by a discussion on the integration of physics and DL in energy systems. The PgNN is a class of neural networks that integrate physical knowledge and principles into the network learning process. In PgNNs, the PBM and DL components are designed to function autonomously by executing their tasks without continual explicit interaction. For instance, by incorporating the output of PBM into the DL model as an additional input, the DL model gains prior physical knowledge, thereby guiding its functionality to achieve better results.

In a recent study, Zerrougui et al. [12] proposed a physics-informed neural network with physical laws governing temperature distribution in proton exchange membrane electrolysis, and demonstrated that their method outperformed recurrent models such as LSTM. In another instance, Lee et al. [13] proposed a physics-informed XGBoost model that integrates physical constraints derived from a modified Nusselt model to enhance the prediction of condensation Heat Transfer Coefficients (HTCs). In their setup, the data-driven model was XGBoost, while the physics-based component stemmed from analytical degradation factors embedded in the modified Nusselt model. The proposed method was compared to standalone Machine Learning (ML) methods such as multilayer perceptron and random forest regression. Physics-informed-XGBoost significantly improved extrapolation performance, reducing Mean Absolute Percentage Error (MAPE) to 11.22% on unseen experimental conditions.

The SIRD (Susceptible, Infected, Recovered, and Deceased) is an ordinary differential equations-based hybrid model that helps feed-forward neural network to forecast disease spread by estimating key factors such as contact rates, death rates, and recovery rates. SIRD uses mathematical equations to describe how people move between these categories over time based on variables such as disease transmission, death rate, and recovery rate. The DL model utilizes SIRD output and real-world data to improve the predictions [14]. Cross Correlation-based Simulation (CCSIM) is also a PBM: it enhances datasets for a deep CNN by generating diverse images of porous media to analyze permeability and morphology [15]. $W_{Mod}$ is a PBM that provides realistic rendering of specific phenomena (e.g., raindrops), and the DL-based image-to-image translation Generative Adversarial Net (GAN) learns to generate the complex visual traits of the scene with high photorealistic quality [16]. In another similar study, outputs from PBMs, such as the Princeton ocean model, the hybrid coordinate ocean model, and the finite-volume coastal ocean, were solely used to train GANs. After that, the DL model was fine-tuned with actual data [17]. Chen et al. [18] proposed a multi-agent deep reinforcement learning algorithm that utilizes the physics-based global voltage sensitivity to enhance its training process. Bento et al. [19] proposed a PgNN to compute the load margin of power systems, regularized to reconstruct the power flow equations at the threshold defining the load margin.

In another study, Hofmann et al. [20] proposed a physics-constrained transfer learning approach that combines a hybrid Temporal Convolutional Network–Long Short-Term Memory (TCN-LSTM) model with a mechanistic model based on electrode Open-Circuit Potential (OCP) alignment. The DL-based TCN-LSTM model estimates the Open-Circuit Voltage (OCV) curve or alignment parameters from partial charging segments, while the physics-based model constrains the output by reconstructing the OCV via optimization or analytical equations. The system achieved Mean Absolute Errors (MAEs) below 10 mV for OCV reconstruction and mean absolute percentage errors under 2% for state-of-health estimation, establishing excellent accuracy and data efficiency compared to purely DL or physics-only baselines.

Nutkiewicz et al. [21] introduced a Data-driven Urban Energy Simulation (DUE-S) that combined physics-based simulations from EnergyPlus with a deep learning residual network (ResNet), where simulated time-series data-encoding thermal dynamics were used as inputs to capture nonlinear urban interactions via DL. Their results showed a Coefficient of Variation of Root Mean Square Error (CV-RMSE) of 0.460 for hourly building-scale predictions and 0.256 for urban-scale predictions, with an additional finding that urban context consideration could enhance retrofit energy savings by up to 7.4%. In another study, Chen et al. [18] proposed a physics-informed neural network model for building thermal modeling and demand response: they embedded physical constraints from an resistance–capacitance model into the neural network's loss function, ensuring predictions adhered to thermal dynamics. They integrate physics with DL to penalize deviations from expected behavior. The model achieved an MAE of 0.25 °C and a CV-RMSE of 1.2% for room temperature, alongside an MAE of 110 W and CV-RMSE of 17.1% for cooling load, outperforming conventional neural networks in accuracy and physical consistency.

Xiao et al. [22] presented a physics-informed recurrent neural network structure to enhance building thermal modeling and energy optimization. They augmented an LSTM network embedding physical constraints such as the directional impact of heating and cooling on temperature through positive definite weight matrices, merging physical principles with temporal modeling capabilities of DL. Their hybrid recurrent network-based controller yielded energy savings of 5.8% compared to On/Off controllers, 4.5% versus state–space models, and 8.9% over LSTM-based controllers while improving thermal comfort by 55%, 59%, and 64%, respectively.

In another study, Tian et al. [23] devised an EnergyPlus-GAN (E-GAN) model to forecast power demand across large-scale building populations. They utilized EnergyPlus simulations of a representative building subset to provide physics-informed inputs, which were then processed by a generative adversarial network, blending physical simulation with the generative power of DL to extrapolate demand patterns. The E-GAN reduced MAPE by approximately 70% compared to traditional data-driven models such as support vector machine.

Similarly, Ma et al. [24] proposed a physics-informed ensemble learning framework incorporating residual modeling for building thermal load prediction. This method integrates EnergyPlus simulation outputs with LSTM, using physics-based data for deterministic components and DL to model residuals reflecting occupant-driven stochasticity. The ensemble approach improved prediction accuracy by 40%–90% in MAE and CV-RMSE over purely physics-based models, achieving CV-RMSE values of 0.300 for cooling and 0.172 for heating with just 10% of annual training data, significantly outperforming a standalone LSTM. Brøgger et al. [25] proposed a hybrid modeling approach to estimate heating energy consumption in building stocks by integrating physics-based simulation data with machine learning. They combined the estimated energy demand from the European ISO 13970 standard with observed data, such as energy use and building characteristics, and passed these into a multiple linear regression model to leverage physical insights alongside data-driven techniques. Their results showed a 13.3% reduction in the CV-RMSE compared to purely physics-based models, demonstrating improved accuracy in energy predictions.

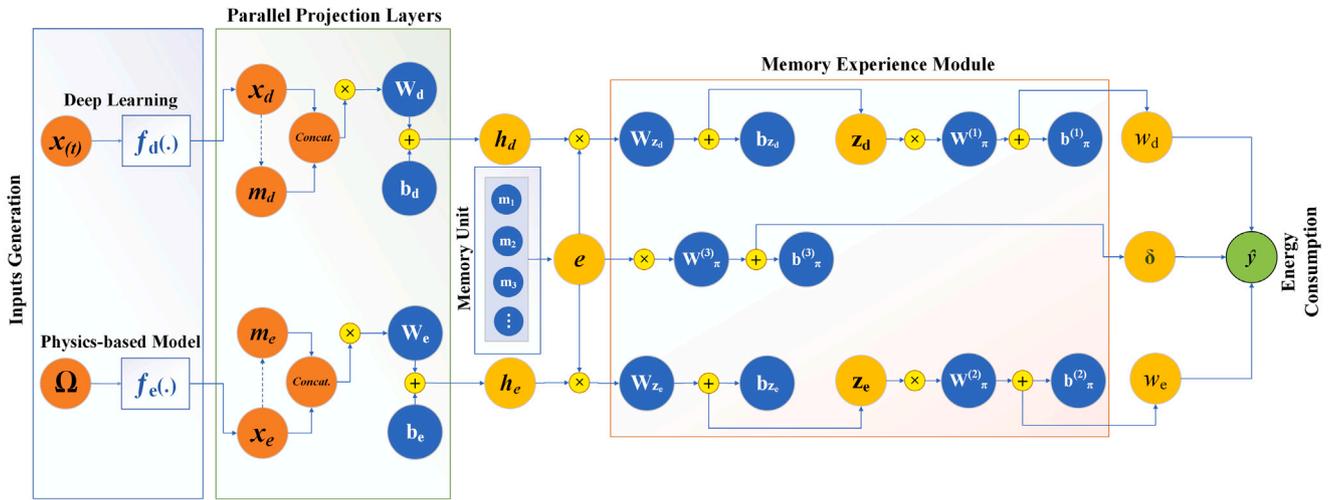

**Fig. 1.** The Physics-Guided Memory Network architecture comprises four primary components: Input Generations, Parallel Projection Layers, Memory Unit, and Memory Experience Module. The Parallel Projection Layers transform input features into learned representations, while the Memory Unit captures persistent biases, and the Memory Experience Module optimally combines inputs to produce accurate energy forecasts. The Memory Unit is a learnable vector across training instances that captures forecasting biases.

In another study, Jiang et al. [26] introduced a Modularized Neural Network (ModNN) incorporating physical priors for building energy modeling, focusing on indoor temperature prediction. This approach decomposed heat balance equations into distinct neural network modules, each estimating specific heat transfer terms, and utilized a sequence-to-sequence encoder–decoder structure with gated recurrent units to blend physical knowledge with temporal modeling capabilities of DL. The ModNN achieved an average MAE of 0.43 °C and an MAPE of 1.93%, and showed greater robustness than a standalone LSTM model across varying training data sizes.

Although various studies have begun integrating physics-based and DL models in the energy domain, they either use EnergyPlus outputs as additional inputs to DL models [21], embed physical laws as constraints into the model structure [18,22], or use simulation results for data augmentation or ensemble calibration [11]. Despite their successful deployments, these hybrid approaches remain limited to specific use cases, such as improving accuracy or incorporating physics as a static input rather than addressing the broader range of scenarios encountered in real-world energy systems. In particular, their applicability is limited in diverse real-world situations where standalone DL or physics-based models fail, such as in newly constructed buildings (where historical data is absent and DL models cannot learn) or in scenarios with missing or sparse data (where DL performance degrades due to missing data). In contrast, our paper merges DL and BEMs not merely for performance enhancement but to enhance applicability in diverse real-world situations where standalone DL or BEM cannot be applied.

### 2.2. Building energy modeling simulation tools

EnergyPlus is a widely adopted building energy modeling tool that simulates thermal, electrical, and mechanical energy performance in buildings. As a next-generation building energy modeling system built upon the load algorithms of the Basic Local Alignment Search Tool (BLAST) and the system algorithms of DOE-2, EnergyPlus analyzes and estimates energy consumption under different configurations and control strategies. It takes Input Data Files (IDF) containing building geometry, surfaces, zones, thermal properties of building materials, HVAC systems specifications, lighting, occupancy patterns, and weather data to predict energy consumption. Unlike the standalone sequential engine of DOE-2, EnergyPlus runs all algorithms and computations simultaneously [27,28]. EnergyPlus can be used with third-party graphical user interfaces such as OpenStudio, a cross-platform software tool designed to make EnergyPlus more accessible. It provides a user-friendly interface, graphical input, and visualization tools to support whole-building energy modeling and advanced energy system analysis [29].

DesignBuilder is another widely used interface for EnergyPlus. It provides a graphical environment for creating and editing input data files and streamlining building simulations [30].

However, discrepancies exist in simulation results between different PBMs, which can be attributed to differences in parametrizations and modeling approaches [28]. Despite this, predictions from PBMs follow a specific pattern that DL models can utilize. For instance, when a sudden change in the building's infrastructure, such as replacing the HVAC system, EnergyPlus and other PBMs can update their predictions based on the new building model. In contrast, if a DL model is trained on the same building, it learns solely from historical data and may produce inaccuracies when the HVAC system is updated. Therefore, providing outputs from PBMs to DL models can yield more reliable results [28,31]. Table 1 summarizes the characteristics of both systems across different aspects.

Although BEMs provide physically grounded simulations of energy consumption, their integration with DL models for improvement of energy forecasting remains largely unexplored. This paper addresses this opportunity by proposing a Physics-Guided Memory Network. The proposed PgMN addresses several challenges, including energy forecasting for new buildings, handling incomplete training data, and managing sparse or missing samples. The paper also investigates how incorporating physics-based simulations enables the network to align its predictions with physically simulated energy patterns, improving overall accuracy and robustness.

## 3. Physics Guided Memory Network

This section describes data and variable representation, physics-based energy simulation and DL forecasts, Parallel Projection Layers, Memory Unit, Memory Experience Module, end-to-end backpropagation, and model scalability. The architecture is illustrated in Fig. 1, and details are given in the following subsections. The detailed functionality of the Physics-Guided Memory Network is outlined in Algorithm 1, which describes the end-to-end workflow of the network.

### 3.1. Data and variable representation

We consider two energy modeling systems: $f_d$ denotes a DL model (e.g., LSTM), and $f_e$ denotes a Physics-Based model (e.g., EnergyPlus).

**Table 1**
Comparison of deep learning and physics-based models.

| Aspect | Deep learning models | Physics-based models |
| --- | --- | --- |
| Data Requirements | Needs substantial historical data | Requires detailed building parameters |
| Modeling Approach | Deterministic or Probabilistic methods | Based on physical laws |
| Occupant Behavior | Learns occupant patterns implicitly | Must encode user schedules explicitly |
| New Construction | Not usable if no past data | Useful even without historical data |
| Modeling Time | Faster setup if data is prepared | Time-intensive building parameter setup |
| Response to Shifts | May fail if new changes are not trained | Adapts if changes align with physics |
| Interpretability | Often viewed as a "black box" | Transparent, based on physics equations |

**Table 2**
Situations addressed by PgMN to improve forecasting applicability.

| Input availability scenarios | Binary mask setting | Imputation strategy | Forecasting behavior and model response |
| --- | --- | --- | --- |
| *Partially Missing DL Prediction* | $m_d = 0$ (where imputation not possible), $m_e = 1$ | Mean of neighbor values or 0 if not possible | PgMN leverages DL (after impute) and available simulated data $x_e$ to support learning. |
| *Fully Missing DL Prediction* | $m_d = 0$ (entirely), $m_e = 1$ | DL inputs set to zero | PgMN learns solely from physically simulated $x_e$ to produce predictions and remains functional for new buildings. |
| *Partially Missing EP Simulated Energy* | Not applicable | Not applicable | EnergyPlus generates complete simulation energy; partial missing data for $x_e$ does not occur in practice. |
| *Fully Missing EP Simulated Energy* | $m_e = 0$, $m_d = 1$ | EP inputs set to zero | PgMN operates using only DL predictions, where EnergyPlus is unavailable or impractical. |
| *Both Inputs Available* | $m_d = 1$, $m_e = 1$ | Not applicable | PgMN learns from both DL and EP-simulated energy and produces predictions by minimizing loss across both sources. |

The DL model takes input features, symbolized as $x_t$ (such as temperature, previous energy consumption, and time), and forecasts energy consumption expressed as $x_d \in \mathbb{R}$. The physics-based model takes building parameters denoted as $\Omega$ (such as HVAC, wall thickness, and insulation properties) and produces simulated energy represented as $x_e \in \mathbb{R}$.

The PgMN is designed to handle situations summarized in Table 2. If the prediction data from either $f_d$ or $f_e$ are missing due to any reason, we treat the corresponded data in the affected model as missing data. To handle missing data, we introduced binary masks $m_d, m_e \in \{0, 1\}$ to enable PgMN to comprehend which data are missing or present. This is equivalent to parts of the data missing, but it assists in processing the data. A mask $m_d = 1$ indicates that $x_d$ is available, and $m_d = 0$ otherwise. Similarly, $m_e = 1$ if $x_e$ is available, and $m_e = 0$ otherwise. In practice, EnergyPlus generates simulated energy for every time step and is unlikely to have missing data. When an energy prediction is unavailable $m_d = 0$ the corresponding $f_d$ input is set to the mean of its neighbor values or 0 if no neighbors are available, without affecting $f_e$. The ground-truth energy consumption is denoted by $y \in \mathbb{R}$, and the final predicted energy value is denoted as $\hat{y} \in \mathbb{R}$. The loss function minimized during training is expressed as $\mathcal{L}(y, \hat{y})$. For time-series data, the time index is denoted by $t$, and training samples are indexed as $n = 1, \ldots, N$. The building parameters, HVAC, wall thickness, and such, are denoted using $\Omega$.

*3.2. Input generations with physical simulated energy and data-driven forecasts*

We require two sets of forecasts, $x_e$ from a physics-based model and $x_d$ from a DL model, so that PgMN can optimally integrate physically simulated and data-driven predictions. To produce $x_e$, a physics-based model (EnergyPlus) solves thermodynamic and heat transfer equations over building parameters $\Omega$, which can be wall thickness, ventilation specifics, or similar. This modeling process can be expressed as:

$$x_e = f_e(\Omega_1, \ldots, \Omega_n). \tag{1}$$

To produce $x_d$, a DL model is trained on temporal data $x(t)$, which are prepared using a sliding window technique [32]. The DL model takes the previous 24 h of temporal input features $x_t$, which consist of hourly recorded outdoor temperature, day of month, day of year, day of week, hour, and the energy consumption of the previous time hours. The model aims to forecast future energy $x_d$ for the next $t_n$ time steps– this is expressed as:

$$x_d = f_d(x_{t_1}, \ldots, x t_n), \tag{2}$$

Both $x_e$ and $x_d$ serve as inputs to PgMN, enabling leveraging physically informed and data-driven predictions that help PgMN establish the optimal balance between the two forecasts to minimize the loss and improve performance. While a unified feature fusion for physics and DL streams could be employed, PgMN deliberately adopts separate processing of $x_d$ and $x_e$ through Parallel Projection Layers to ensure the model remains operational even if one source is missing, partially available, or unreliable, which are common challenges in real-world building environments. The separation preserves the individual strengths of data-driven and physics-based predictions, avoids early information mixing, and allows the Memory Experience Module to learn an optimal combination later, flexibly and robustness.

## 3.3. Parallel Projection Layers

The purpose of the Parallel Projection Layers is to map the input scalar predictions $x_d$ and $x_e$, along with their corresponding binary masks $m_d$ and $m_e$, into learned deep features: this is represented in lines 11 to 13 in Algorithm 1. The masks $m_d$ and $m_e$ are prepared to indicate the availability of $x_d$ and $x_e$, with $m = 1$ when data is present and $m = 0$ otherwise. Additionally, this layer ensures that PgMN can dynamically handle missing data while enhancing applicability across varying scenarios. For the DL prediction $x_d$ and its mask $m_d$, the embedding is computed as:

$$\mathbf{h}_d = \sigma\left(\mathbf{W}_d \begin{bmatrix} x_d \\ m_d \end{bmatrix} + \mathbf{b}_d\right) \tag{3}$$

where $\mathbf{W}_d \in \mathbb{R}^{d \times 2}$ is a learnable weight matrix, $\mathbf{b}_d \in \mathbb{R}^d$ is a learnable bias vector, and the activation function $\sigma(\cdot)$ is a standard Rectified Linear Unit (ReLU) [33]. Similarly, for the EnergyPlus simulated energy consumption $x_e$ and its mask $m_e$, the embedding is given by:

$$\mathbf{h}_e = \sigma\left(\mathbf{W}_e \begin{bmatrix} x_e \\ m_e \end{bmatrix} + \mathbf{b}_e\right) \tag{4}$$

where $\mathbf{W}_e \in \mathbb{R}^{d \times 2}$ and $\mathbf{b}_e \in \mathbb{R}^d$ are learnable parameters. The outputs $\mathbf{h}_d \in \mathbb{R}^d$ and $\mathbf{h}_e \in \mathbb{R}^d$ are intermediate representations or deep features of energy predictions along with their availability. These embeddings are used in subsequent PgMN layers to process the information from both sources and help the model to learn time-varying patterns more easily as they are now transformed into deep features, which is easier to learn compared to raw inputs [34]

Parallel Projection Layers inform the model whether an entire dataset or specific values are missing. This helps the model understand the available data and adjust its processing accordingly. By doing so, the model becomes better at handling incomplete data and can focus on making predictions without being constrained by missing inputs. This method also allows the model to make predictions over an extended range, starting from the available data and continuing beyond, even when some input parts are missing.

Parallel Projection Layers define continuous and differentiable mappings from $\mathbb{R}^2$ (input scalar and mask) to $\mathbb{R}^d$ (the embedding space). As per Universal Approximation Theorem (UAT) [35], such mappings can approximate any continuous function on a compact domain given sufficient hidden units. This property ensures that Parallel Projection Layers are capable of learning complex transformations of the input energy forecasts $x_d$ and $x_e$ while also capturing the effect of their availability as indicated by the binary masks $m_d$ and $m_e$.

## 3.4. Memory Unit (MU)

The Memory Unit (MU) is designed to learn experience from historical data, which helps to reduce biases in energy forecasts, enabling the model to correct errors dynamically. The MU operation js is handled in lines 1, 14, 15 in Algorithm 1. This is a learnable global repository of experiences, such as actual versus predicted energy, that records all previous instances. The memory is represented as a single vector $\mathbf{m} \in \mathbb{R}^{d_m}$, where $d_m$ is dimensionality. Unlike recurrent mechanisms that update over time steps, the Memory Unit operated as a globally shared vector across training instances, capturing persistent forecasting biases and historical correction patterns in a learnable form.

This MU vector is initialized randomly and is updated iteratively during training through backpropagation. The memory output used in the model, denoted as $\mathbf{e}$, is retrieved as:

$$\mathbf{e} = \begin{bmatrix} m_1 \\ m_2 \\ \vdots \\ m_{d_m} \end{bmatrix}, \quad \mathbf{m} \in \mathbb{R}^{d_m} \tag{5}$$

where $\mathbf{m}$ provides a compact representation of persistent errors across all samples. The memory serves as a bias correction term that adjusts the predictions $\hat{y}$ to better align with the ground truth $y$. The memory parameters $\mathbf{m}$ are learned along with other model parameters to minimize the forecasting error. The MU adds a representational capacity to the model, allowing it to adjust systematic biases in the predictions. The memory vector $\mathbf{m}$ is a learnable parameter and differentiable and updated using backpropagation. The gradient for the memory vector is calculated as:

$$\frac{\partial \mathcal{L}}{\partial \mathbf{m}} = \frac{\partial \mathcal{L}}{\partial \mathbf{e}} \cdot \frac{\partial \mathbf{e}}{\partial \mathbf{m}} \tag{6}$$

where $\mathbf{e}$ represents the memory output. This allows the memory to store global corrections that persist across the entire dataset, ensuring it is aligned to minimize forecasting errors. The memory in PgMN serves as a trainable look-up vector that explicitly stores global bias corrections.

## 3.5. Memory Experience Module (MEM)

The Memory Experience Module determines the best balance between EnergyPlus and DL predictions, $x_d, x_e \in \mathbb{R}$, and predicts the output where the loss is minimized, as represented in lines 16 to 24 in the Algorithm 1. If the model cannot find a point between these predictions that minimizes the loss, it allows the prediction to go beyond the range defined by EnergyPlus and DL predictions. The MEM takes the projected embeddings $\mathbf{h}_d \in \mathbb{R}^d$ and $\mathbf{h}_e \in \mathbb{R}^d$ from the Parallel Projection Layers. These embeddings assist in identifying missing values in the predictions. It also utilizes the memory vector $\mathbf{e} \in \mathbb{R}^{d_m}$, which functions as a learnable unbounded weight matrix and incorporates previous experiences. The purpose is to predict final energy prediction $\hat{y}$, allowing the model to exceed the minimum or maximum of $\{x_d, x_e\}$ whenever this reduces the training loss. The hidden representation for $w_d$ is learned by leveraging the memory unit $e$ and the activation $h_d$, which is expressed as:

$$z_d = \sigma\left(\mathbf{W}_{z_d} \cdot \begin{bmatrix} \mathbf{h}_d \\ e \end{bmatrix} + \mathbf{b}_{z_d}\right), \tag{7}$$

where $\sigma(\cdot)$ represents the ReLU activation function, and $\mathbf{W}_{z_d}$ and $b_{z_d}$ are learnable parameters. This Eq. (7) ensures that both the data-driven features and global corrections from memory are learned. Then $z_d$ is passed through a linear transformation without an activation function to compute $w_d$:

$$w_d = \mathbf{W}_\pi^{(1)} \cdot z_d + \mathbf{b}_\pi^{(1)}, \tag{8}$$

The additional layer from Eq. (8) enhances the learning capacity of the model by increasing its depth and allowing the network to express more complex feature interactions. Simonyan et al. [36] showed that deeper architectures, achieved by stacking even simple linear layers, enable the model to capture increasingly abstract and high-level representations.

Similarly, for the EnergyPlus path, the hidden representation $z_e$ is learned by combining the memory unit $e$ with the input $h_e$, as follows:

$$z_e = \sigma\left(\mathbf{W}_{z_e} \cdot \begin{bmatrix} \mathbf{h}_e \\ e \end{bmatrix} + \mathbf{b}_{z_e}\right), \tag{9}$$

Then $z_e$ is passed through a linear transformation without activation to compute $w_e$:

$$w_e = \mathbf{W}_\pi^{(2)} \cdot z_e + \mathbf{b}_\pi^{(2)}. \tag{10}$$

The hidden representation for $\delta$ can be expressed by using a memory unit as:

$$\delta = \mathbf{W}_\pi^{(3)} \cdot [e] + \mathbf{b}_\pi^{(3)}, \tag{11}$$

The parameters $\mathbf{W}_{z_d}, \mathbf{W}_{z_e} \in \mathbb{R}^{d_z \times (d+d_m)}$, $\mathbf{W}_\pi^{(1)}, \mathbf{W}_\pi^{(2)}, \mathbf{W}_\pi^{(3)} \in \mathbb{R}^{d_z}$, and $\mathbf{b}_{z_d}, \mathbf{b}_{z_e}, \mathbf{b}_\pi^{(1)}, \mathbf{b}_\pi^{(2)}, \mathbf{b}_\pi^{(3)} \in \mathbb{R}$ are learnable parameters. The $w_d$ and $w_e$ are outputs that give the PgMN representation of their respective forecasts, and $\delta$ gives the option to move outside of the margin if there is no place between $w_d$ and $w_e$ where the error is minimized. This learnable mechanism ensures that model finds most suitable prediction. The final predicted energy $\hat{y}$ is given by:

$$\hat{y} = w_d + w_e + \delta. \tag{12}$$

If both inputs are present but under or over-predict the energy, the learned parameters can adjust to produce predictions that exceed the numerical range of $\{x_d, x_e\}$, fixing any shared bias. When one input is missing, its absence is captured through the mask information embedded in $\mathbf{h}_d$ and $\mathbf{h}_e$, allowing the corresponding weight to be effectively reduced to near zero. Since $w_d$ and $w_e$ are unbounded, the model can generate predictions beyond the range of the given inputs whenever this helps minimize the loss, $\mathcal{L}(y, \hat{y})$. This enables the model to find the optimal balance between the two forecasts or extend beyond their range to achieve higher accuracy. The binary mask inputs $m_d$ and $m_e$ inform the model in real-time about the presence or absence of data (e.g., a sensor dropout). The Parallel Projection Layers and Memory Experience Module enable the network to remain fully operational by adaptively rebalancing between DL and EnergyPlus inputs, even when one source is entirely missing, ensuring robust predictions under sensor failure or incomplete data scenarios.

### 3.6. End-to-end backpropagation

The PgMN learnable parameters including $(\mathbf{W}_d, \mathbf{b}_d, \mathbf{W}_e, \mathbf{b}_e)$, memory vector $\mathbf{m}$, and memory experience parameters $(\mathbf{W}_{z_d}, \mathbf{W}_{z_e}, \mathbf{b}_{z_d}, \mathbf{b}_{z_e})$ and $(\mathbf{W}_\pi^{(1)}, \mathbf{W}_\pi^{(2)}, \mathbf{W}_\pi^{(3)}, \mathbf{b}_\pi^{(1)}, \mathbf{b}_\pi^{(2)}, \mathbf{b}_\pi^{(3)})$, are updated by backpropagation in the fully differentiable manner aiming to minimize loss function given by: $\mathcal{L}(y^{(n)}, \hat{y}^{(n)})$ measures the difference between $\hat{y}^{(n)}$ and the ground truth $y^{(n)}$. Here, a common Mean-Squared Error (MSE) is employed:

$$\mathcal{L}(y^{(n)}, \hat{y}^{(n)}) = (y^{(n)} - \hat{y}^{(n)})^2. \tag{13}$$

Summing over all samples

$$\min_{params} \sum_{n=1}^{N} \mathcal{L}(y^{(n)}, \hat{y}^{(n)}),$$

drives the end-to-end backpropagation. This follows standard backpropagation principles, but it updates the varying nature of learnable parameters. The standard optimizers (e.g., SGD or Adam) iteratively adjust each parameter to minimize total loss.

## 4. Theoretical evaluation

This section provides theoretical guarantees for the PgMN to ensure its applicability across various scenarios.

### 4.1. Universal function approximation

This subsection aims to ensure that the use of parallel projection layers, memory unit, and their learnable aggregation using the Memory Experience Module still satisfies the conditions of the Universal Approximation Theorem (UAT). We thus establish the theoretical grounding for the practical deployment of PgMN.

**Theorem 4.1** (*Universal Function Approximation by PgMN*). *Let $\mathcal{D} \subset \mathbb{R}^2$ be a compact domain of input forecasts $(x_d, x_e)$, and let $\mathcal{F} : \mathcal{D} \to \mathbb{R}$ be any continuous target function. Suppose PgMN uses parallel projection mappings $\Phi_d : \mathbb{R}^2 \to \mathbb{R}^d$ and $\Phi_e : \mathbb{R}^2 \to \mathbb{R}^d$, a memory vector $\mathbf{m} \in \mathbb{R}^{d_m}$, and a revised aggregator that learns three scalars $(w_d, w_e, \delta)$ and outputs*

$$\hat{y} = w_d + w_e + \delta. \tag{14}$$

*Then for any $\epsilon > 0$, there exist learnable parameters*

$\mathbf{W}_d, \mathbf{b}_d, \mathbf{W}_e, \mathbf{b}_e, \mathbf{m}, \mathbf{W}_{z_d}, \mathbf{b}_{z_d}, \mathbf{W}_{z_e}, \mathbf{b}_{z_e}, \mathbf{W}_\pi^{(1)}, \mathbf{W}_\pi^{(2)}, \mathbf{W}_\pi^{(3)}, b_\pi^{(1)}, b_\pi^{(2)}, b_\pi^{(3)}$

**Algorithm 1** Physics-Guided Memory Network

**Require:**
- Deep Learning forecasts $x_d$,
- EnergyPlus outputs $x_e$,
- binary masks $m_d, m_e$,
- either actual $y$ **or** no actual data available.

**Ensure:**
- Final energy prediction $\hat{y}$

1: **Initialize** trainable memory vector $\mathbf{m} \leftarrow \mathbf{0}$
2: **Initialize** model parameters in:
   - Parallel Projection layers $(\mathbf{W}_d, \mathbf{b}_d)$ and $(\mathbf{W}_e, \mathbf{b}_e)$
   - Memory Experience Module $(\mathbf{W}_{z_d}, \mathbf{W}_{z_e}, \mathbf{b}_{z_d}, \mathbf{b}_{z_e})$ and $(\mathbf{W}_\pi^{(1)}, \mathbf{W}_\pi^{(2)}, \mathbf{W}_\pi^{(3)}, \mathbf{b}_\pi^{(1)}, \mathbf{b}_\pi^{(2)}, \mathbf{b}_\pi^{(3)})$
3: **Initialize** optimizer (e.g., Adam) with learning rate $\eta$
4: **for** each training epoch **do**
5:   **Receive** training samples: $\{(x_d^{(n)}, m_d^{(n)}, x_e^{(n)}, m_e^{(n)}, y^{(n)})\}$ *(or no-y if none)*
6:   **for** each sample $n$ **do**
7:     **if** *Actual data is absent* **then**
8:       $y^{(n)} \leftarrow x_e^{(n)}$    {Use EP as proxy ground truth}
9:     **end if**
10:     **Forward pass:**
11:     Compute activations from Parallel Projection Layers:
12:     $\mathbf{h}_d \leftarrow \sigma\left(\mathbf{W}_d [x_d^{(n)}, m_d^{(n)}]^T + \mathbf{b}_d\right)$
13:     $\mathbf{h}_e \leftarrow \sigma\left(\mathbf{W}_e [x_e^{(n)}, m_e^{(n)}]^T + \mathbf{b}_e\right)$
14:     Retrieve memory vector:
15:     $e \leftarrow \mathbf{m}$
16:     Compute hidden representations:
17:     $z_d \leftarrow \sigma\left(\mathbf{W}_{z_d} \cdot \begin{bmatrix}\mathbf{h}_d \\ e\end{bmatrix} + \mathbf{b}_{z_d}\right)$
18:     $z_e \leftarrow \sigma\left(\mathbf{W}_{z_e} \cdot \begin{bmatrix}\mathbf{h}_e \\ e\end{bmatrix} + \mathbf{b}_{z_e}\right)$
19:     Compute weights and offset:
20:     $w_d \leftarrow \mathbf{W}_\pi^{(1)} \cdot z_d + \mathbf{b}_\pi^{(1)}$
21:     $w_e \leftarrow \mathbf{W}_\pi^{(2)} \cdot z_e + \mathbf{b}_\pi^{(2)}$
22:     $\delta \leftarrow \mathbf{W}_\pi^{(3)} \cdot e + \mathbf{b}_\pi^{(3)}$
23:     Compute final prediction:
24:     $\hat{y}^{(n)} \leftarrow w_d + w_e + \delta$
25:     **Compute** loss:
26:     $\mathcal{L}(y^{(n)}, \hat{y}^{(n)}) \leftarrow (y^{(n)} - \hat{y}^{(n)})^2$
27:     **Calculate** gradients w.r.t. all parameters and memory vector $e$
28:   **end for**
29:   **Update** model parameters and memory vector $\mathbf{m}$ using gradient descent with rate $\eta$    {e.g., **param** ← **param** − $\eta \nabla \mathcal{L}$}
30: **end for**
31: **After training**, to predict for new inputs $(x_d, m_d, x_e, m_e)$:
32: **Use** final parameters and memory vector $\mathbf{m}$ to compute:
33: $\mathbf{h}_d, \mathbf{h}_e, z_d, z_e, w_d, w_e, \delta \to \hat{y} = w_d + w_e + \delta$
34: **return** $\hat{y}$

*such that*

$$\sup_{(x_d, x_e) \in \mathcal{D}} \left| \mathcal{F}(x_d, x_e) - \hat{y}(x_d, x_e) \right| < \epsilon, \tag{15}$$

*where $\hat{y}(x_d, x_e)$ is given by applying parallel projections $\Phi_d, \Phi_e$ and the memory-based aggregator that computes $w_d, w_e, \delta$ from $\mathbf{h}_d, \mathbf{h}_e, \mathbf{m}$, then summing them as in Eq. (14). The supremum (sup) is the least upper bound of a function's values over a given domain.*

**Proof.** Let $(x_d, m_d) \mapsto \mathbf{h}_d \in \mathbb{R}^d$ and $(x_e, m_e) \mapsto \mathbf{h}_e \in \mathbb{R}^d$ be the parallel projection mappings. By the Universal Approximation Theorem, each projection layer can approximate any continuous function on a compact set if it is sufficiently wide or deep. Hence, for appropriate choices of $\mathbf{W}_d, \mathbf{b}_d$ and $\mathbf{W}_e, \mathbf{b}_e$, the pair $(\mathbf{h}_d, \mathbf{h}_e)$ can embed $(x_d, x_e)$ into $\mathbb{R}^{2d}$ in a manner that captures the essential structure of the desired mapping $\mathcal{F}$.

Let $\mathbf{m} \in \mathbb{R}^{d_m}$ be the memory vector that is added to each embedding path or used independently to produce $\delta$. We define three separate streams for $(\mathbf{h}_d, \mathbf{m}) \mapsto w_d$, $(\mathbf{h}_e, \mathbf{m}) \mapsto w_e$, and $\mathbf{m} \mapsto \delta$. Each of these streams is effectively a feed-forward map $\mathbb{R}^{d+d_m} \to \mathbb{R}$ (or $\mathbb{R}^{d_m} \to \mathbb{R}$ for $\delta$) that the UAT also guarantees can approximate arbitrary continuous mappings on compact domains. Summing the partial outputs $w_d + w_e + \delta$ then yields $\hat{y}$, as in Eq. (14). Hence, the composition
$$(x_d, x_e) \longmapsto (\mathbf{h}_d, \mathbf{h}_e, \mathbf{m}) \longmapsto (w_d, w_e, \delta) \longmapsto \hat{y}(x_d, x_e) = w_d + w_e + \delta \tag{16}$$

can approximate $\mathcal{F}(x_d, x_e)$ to within any $\varepsilon > 0$, provided $d, d_m, d_z$ (and associated hidden units) are chosen large enough. □

This theorem confirms that, under mild assumptions (Differentiability, bounded and closed set (compact domain)), PgMN can learn any continuous relationship between the two forecast inputs $(x_d, x_e)$ and the desired output $y$.

### 4.2. Bias Correction Capability of the Memory Unit

This subsection shows that the Memory Unit can reduce systematic biases in forecasts by adjusting the learnable vector $\mathbf{m}$. We formalize it via the following statement:

**Theorem 4.2** (*Bias Correction Capability*). *The forecasts $(x_d^{(n)}, x_e^{(n)})$ exhibit systematic bias relative to the true values $y^{(n)}$. Let $\mathbf{m} \in \mathbb{R}^{d_m}$ be PgMN's memory vector, and write $\hat{y}^{(n)}(\mathbf{m})$ for the final output at sample $n$ as given in Eqs. (7)–(12) (i.e., through $w_d, w_e, \delta$). Consider the mean-squared error loss*
$$\mathcal{L}(\mathbf{m}) = \sum_{n=1}^{N} \left( y^{(n)} - \hat{y}^{(n)}(\mathbf{m}) \right)^2. \tag{17}$$

*Then standard gradient-based training iteratively shifts $\mathbf{m}$ to reduce the average bias between $\hat{y}^{(n)}(\mathbf{m})$ and $y^{(n)}$ over the dataset.*

**Proof.** Let $\hat{y}^{(n)}(\mathbf{m})$ denote PgMN's final prediction at sample $n$. By Eqs. (7)–(12), the scalar $\hat{y}^{(n)}(\mathbf{m})$ is a differentiable function of $\mathbf{m}$. The total loss is
$$\mathcal{L}(\mathbf{m}) = \sum_{n=1}^{N} \left( y^{(n)} - \hat{y}^{(n)}(\mathbf{m}) \right)^2. \tag{18}$$

Taking partial derivatives with respect to $\mathbf{m}$ and applying the chain rule yields:
$$\frac{\partial \mathcal{L}}{\partial \mathbf{m}} = \sum_{n=1}^{N} 2 \left( \hat{y}^{(n)}(\mathbf{m}) - y^{(n)} \right) \frac{\partial \hat{y}^{(n)}(\mathbf{m})}{\partial \mathbf{m}}. \tag{19}$$

If the forecasts systematically deviate from $y^{(n)}$ (e.g., persistently underpredict), then $\hat{y}^{(n)}(\mathbf{m}) - y^{(n)}$ has a consistent sign in the above summation. Under a gradient-descent update,
$$\mathbf{m} \leftarrow \mathbf{m} - \eta \, \nabla_{\mathbf{m}} \mathcal{L}(\mathbf{m}), \tag{20}$$

the memory vector moves in a direction that reduces the average bias term over all $n$. Because each $\hat{y}^{(n)}(\mathbf{m})$ depends on $\mathbf{m}$ through the aggregator (Eqs. (7)–(12)), the aggregated gradient can correct systematic errors in either $\mathbf{h}_d$ or $\mathbf{h}_e$. This establishes that $\mathbf{m}$ operates as a persistent bias-correction term and is tuned to compensate for mismatch in $(x_d, x_e)$. □

This shows that the learnable memory vector $\mathbf{m}$ can record and reduce systematic deviations in the two input forecasts. As $\mathbf{m}$ adjusts, the Memory Experience Module shifts $\hat{y}^{(n)}$ to align with true values $y^{(n)}$, even when the raw inputs $(x_d, x_e)$ exhibit consistent bias.

### 4.3. Unbounded output

Since the purpose of PgMN is to establish the best balance between two predictions, such as EnergyPlus and DL, and find the best place where the error is minimized, the question about unbounded output becomes necessary to ensure that the model can go outside the margin of both predictions if remaining strictly between them does not minimize the loss.

**Theorem 4.3** (*Unbounded Output Capability*). *Let*
$$\hat{y} = w_d + w_e + \delta \tag{21}$$

*be PgMN's final output, where $(w_d, w_e, \delta) \in \mathbb{R}^3$ are the scalar values produced by the aggregator's linear map in Eqs. (7)–(12). If $w_d$ and $w_e$ are unconstrained real scalars, then for each sample $(x_d, x_e)$, PgMN can set $\hat{y}$ outside the numeric interval spanned by $\{x_d, x_e\}$ if it lowers the loss $\mathcal{L}(y, \hat{y})$.*

**Proof.** From the aggregator's final step,
$$\hat{y} = w_d + w_e + \delta, \tag{22}$$

where $(w_d, w_e, \delta)$ come from learnable transformations with no constraints on sign or magnitude. Because there are three degrees of freedom $(w_d, w_e, \delta)$ and only one target scalar $\hat{y}$, one can always achieve any real value $\hat{y}^*$. If, for instance, a lower loss occurs at some $\hat{y}^* > \max\{x_d, x_e\}$, gradient-based updates to $\mathbf{W}_{z_d}, \mathbf{W}_{z_e}, \mathbf{b}_{z_d}, \mathbf{b}_{z_e}, \mathbf{W}_\pi, \mathbf{b}_\pi, \mathbf{m}$ will push $w_d + w_e + \delta$ above $\max\{x_d, x_e\}$. Likewise, if a smaller value than $\min\{x_d, x_e\}$ reduces the loss, the model can shift $\hat{y}$ below both inputs. Thus, the aggregator's unbounded real parameters guarantee the ability to produce final predictions outside the original forecast range whenever that is optimal.

Hence, the Memory Experience Module can place $\hat{y}$ beyond either input's numeric range to further reduce $\mathcal{L}(y, \hat{y})$ whenever both inputs $(x_d, x_e)$ are biased or insufficiently low/high for optimal performance. □

This proves that the MEM component can produce predictions beyond the range whenever it is necessary to reduce the loss and ensure that the MEM mechanism is theoretically capable of correcting extreme cases where both forecasts are biased.

### 4.4. PgMN applicability in diverse real world scenarios

The PgMN adapts to different real-world conditions by exploiting its Memory Unit and Memory Experience Module to handle missing or incomplete forecasts or targets. In newly constructed buildings where no historical data exist, actual measurements $y$ are unavailable, and the DL model cannot be trained traditionally. The model sets $x_d = 0$ and treats the physics-based simulated energy $x_e$ as input and proxy targets. The memory unit then learns any persistent biases relative to the simulated data, and the unbounded scalar weights in the MEM allow the output to move beyond or strictly match the EnergyPlus profile, ensuring that the model can reduce error even with only one type of forecast. Moreover, we can also train the PgMN model using physics-based simulated energy, which at least adheres to the physical parameters of buildings.

If actual data are partially missing, PgMN simply zeros out the missing segments and relies on the available valid forecasts. Under mild assumptions (i.e., Theorem 4.2), the memory vector corrects for any systematic offset arising from incomplete data. Meanwhile, the unbounded property (Theorem 4.3) guarantees that the MEM can produce outputs beyond the range of either forecast to reduce the loss further.

A similar scenario variant appears when EnergyPlus outputs are unavailable, forcing PgMN to rely on $x_d$ alone. Setting $x_e = 0$ and $m_e = 0$ signals to the Parallel Projection Layer that the physics-based forecast is missing, and the MEM dynamically adjusts to the DL signals.

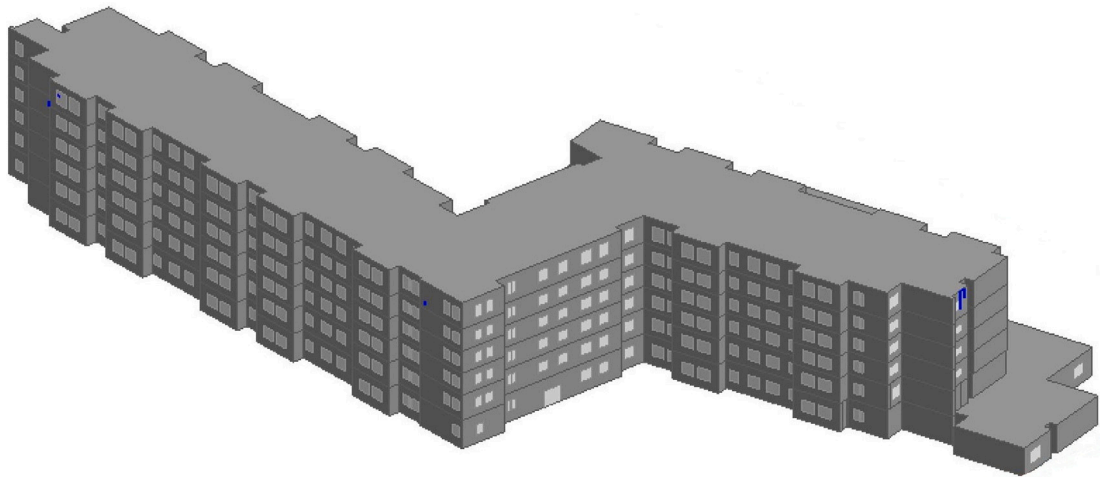

**Fig. 2.** 3D model of the residential building under investigation developed using DesignBuilder for EnergyPlus simulation.

In practice, this is relevant if building parameters $\Omega$ cannot be obtained or if the physics-based model fails to run. In contrast, if both forecasts and actual data are fully present, PgMN benefits from simultaneously learning from physically informed patterns and data-driven predictions. The Memory Unit then stores persistent global corrections while the MEM finds the best mixture of $x_d$ and $x_e$. Under Theorem 4.1, this mixture can universally approximate any target function on the compact domain of forecasts.

These variants show that PgMN's design, including the Memory Unit for bias correction and the Memory Experience Module for unbounded combinations of forecasts, preserves theoretical rigor and practical applicability in each scenario.

## 5. Evaluation and results

This section describes the building selection, evaluation with BEMs (physics-based methods), evaluation with DL models, and evaluation of PGmN.

### 5.1. Building selection

To model a building, we selected a student residence facility in London, Ontario, Canada, which is a building accommodating approximately 495 residences. It has a combination of single and double rooms, with each pair of rooms sharing a semi-private washroom. The building spans six levels, organized into 11 distinct wings, with around 45 residents per floor. The residence also has a range of amenities, including a dedicated dining hall, a cozy lounge with a fireplace, a music practice room. This building is relatively large, but it is still a great choice for examination in both physics-based and DL models.

### 5.2. Evaluation with BEM (EnergyPlus)

We first modeled the entire building in EnergyPlus, version 9.4.0, and employed DesignBuilder, version 7.1.3.015, as a graphical interface to simplify and streamline the modeling process. The modeling began by creating the building's 3D geometry in DesignBuilder, containing architectural details such as floor dimensions, levels, room layouts, and internal partitions to align with the building's design. The construction materials were then assigned to walls, roofs, floors, and windows using the material library in DesignBuilder, specifying insulation levels, thermal conductivity, and glazing attributes to support energy modeling. The building was divided into thermal zones to reflect distinct functional areas such as bedrooms, common lounges, dining spaces, and other areas. We assigned the zone occupancy, ventilation, and temperature control settings based on its specific usage. The local weather data used in EnergyPlus was the Typical Meteorological Year (TMY) data for London, Ontario, sourced from the EnergyPlus Weather (EPW) database. This dataset includes hourly values for key climate variables such as dry-bulb temperature, dew point, solar radiation, wind speed, and direction, representing a statistically typical year of weather patterns for the location. The HVAC systems were specified to include heating, cooling, ventilation, and air distribution configurations. The system operating schedules were configured to match realistic operational performance assumptions based on the Ontario building code. Internal heat gains such as lighting, equipment, and occupant loads were defined for each zone, and detailed schedules of people living in the building were also added to ensure accurate energy consumption prediction. A representative 3D model of the building created in DesignBuilder is provided in Fig. 2 to assist visualization.

The simulation model was calibrated using measured energy consumption data, following the American Society of Heating, Refrigerating and Air-Conditioning Engineers (ASHRAE) Guideline 14-2014 [37]. The calibration accuracy was assessed using two standard metrics recommended by ASHRAE: Normalized Mean Bias Error (NMBE) [37] and the Coefficient of Variation of Root Mean Square Error (CV-RMSE) [37]. The guideline recommends that NMBE and CV-RMSE should not exceed 30% for hourly data and 15% for monthly data. The model was iteratively refined to meet these criteria by adjusting material properties, operational schedules, and especially occupant behavior. Table 4 presents the calibration results, verifying that the simulation meets ASHRAE's recommended accuracy thresholds. Note that measured and simulated energy for the year are the same for hourly and monthly forecasting; however, the errors differ because forecasts at different granularities yield different variations. The modeling requires several processes to complete, which is also given in the EnergyPlus user manual; for our experience with this building, we have added processes for major steps and also provided information that we modeled in our case. The HVAC system was modeled using the `ZoneHVAC:FourPipeFanCoil` object in EnergyPlus, a forced-convection hydronic unit with a supply fan, hot water heating coil, and chilled water cooling capable of heating and cooling through separate water loops. Specifically, the system functioned as a two-pipe fan coil unit, operating in either heating or cooling mode based on the seasonal availability defined in the schedule. This configuration provided space heating, cooling, ventilation, and air distribution while maintaining thermal comfort according to the Ontario Building Code [38]. Table 3 summarizes essential parameters such as occupancy, equipment load, envelope properties, and HVAC set points. The main steps are:

1. **Geometry and Zoning:** We started by defining the floor-by-floor layout in DesignBuilder, assigning distinct zones (e.g., rooms,

**Table 3**
Key building data inputs to EnergyPlus for simulation.

| S.No. | Item | Value |
| --- | --- | --- |
| 1 | Occupancy | Density: 0.10 people/m$^2$ |
| | | Activity: Light office work |
| | | Residents: 495 (approx. 500) |
| 2 | Equipment Load | 20.5 W/m$^2$ |
| | | Radiant fraction: 0.2 |
| 3 | Exterior Wall and Infiltration | Wall: Brick–XPS–Concrete–Gypsum |
| | | U-value: 0.351 W/(m$^2$ K) |
| | | Infiltration: 0.70 ACH (air changes/hour) |
| 4 | Window | Type: Double glass, 6 mm + 13 mm Argon gap |
| | | U-value: 2.5 W/(m$^2$ K) |
| 5 | Lighting | Rooms: 3.3 W/m$^2$ (100 lux) |
| | | Storage/Mech.: 1.8 W/m$^2$ (100 lux) |
| | | Lobby: 1.6 W/m$^2$ (100 lux) |
| 6 | HVAC System | Occupied setpoints: 21–23 °C |
| | | Heating setback: 15 °C, Cooling setback: 28 °C |
| | | Continuous operation with scheduled ventilation |

**Table 4**
Comparison of measured and simulated energy consumption.

| Time interval | Measured (kWh) | Simulated (kWh) | NMBE (%) | CV-RMSE (%) |
| --- | --- | --- | --- | --- |
| Hourly | 2,106,419.0 | 2,064,623.9 | −1.94 | 20.7 |
| Monthly | 2,106,419.0 | 2,064,623.9 | 2.02 | 9.6 |

lobby, storage) according to function and occupancy. This zoning ensures different parts of the building can have individualized operational schedules.

2. **Construction and Envelope:** Then we specified wall assemblies, roof and floor details, window configurations, and infiltration rates according to Canadian building codes and ASHRAE standards [39]. This includes setting the appropriate U-values, as listed in Table 3.

3. **Internal Loads:** Next occupant density, equipment loads, and lighting power densities were specified. These values (e.g., 20.5 W/m$^2$ for equipment) guide the internal heat gains responsible for heating and cooling demand throughout the building. This is modeled as per the Ontario Building Code.

4. **Occupant Schedules and Behavior:** Since this is an undergraduate residence, carefully collected weekday/weekend and seasonal schedules of individuals capture variability in building usage, for instance. These schedules impact lighting usage, equipment operations, and HVAC demands.

5. **HVAC Configuration:** Temperature set points for occupied and setback periods (21–23 °C vs. 15–28 °C) were assigned. The HVAC system in this model runs continuously, but ventilation rates can be programmed to fluctuate with occupancy levels.

6. **Export to EnergyPlus:** Once the construction, loads, and system details were finalized in DesignBuilder, the IDF (Input Data File) was exported into EnergyPlus.

7. **Simulation and Analysis:** In the last step, we executed EnergyPlus runs to evaluate energy consumption and system performance.

The simulated energy consumption pattern produced by EnergyPlus for the entire building is shown in Fig. 3.

### 5.3. Evaluation with deep learning

We examined three DL models for energy prediction, including Transformer, LSTM, and GRU, using their standard architectures [32, 40]. The hyperparameters of each model were tuned utilizing GridSearch [32]: parameter ranges, and the optimal hyperparameters for each model are provided in Table 5. The three evaluation metrics were used in this analysis to ensure fair assessment:

- **Mean Absolute Error (MAE):** Measures the average magnitude of errors in predictions without considering their direction:

$$\text{MAE} = \frac{1}{N}\sum_{i=1}^{N} |y_i - \hat{y}_i|. \quad (23)$$

- **Root Mean Squared Error (RMSE):** Provides a quadratic mean of errors, giving more weight to more significant errors:

$$\text{RMSE} = \sqrt{\frac{1}{N}\sum_{i=1}^{N} (y_i - \hat{y}_i)^2}. \quad (24)$$

- **Symmetric Mean Absolute Percentage Error (SMAPE):** A scale-independent metric for percentage-based error:

$$\text{SMAPE} = \frac{1}{N}\sum_{i=1}^{N} \frac{|y_i - \hat{y}_i|}{\frac{|y_i|+|\hat{y}_i|}{2}} \times 100. \quad (25)$$

The performance of the DL models is summarized in Table 6, which reports the model performance on the testing period with hourly energy consumption where LSTM shows slightly better performance with an SMAPE of 10.012, MAE of 23.490, and RMSE of 26.573. Although all these DL models performed comparably well, we selected LSTM predictions for further analysis due to their slightly better performance. The results are also presented in Fig. 3, showing actual vs predicted values produced by DL and EnergyPlus models.

### 5.4. Evaluation of physics guided memory network

This subsection describes the results and evaluation of our proposed PGmN under five different scenarios.

#### 5.4.1. Scenario 1: EnergyPlus and deep learning predictions available

In many operational buildings, EnergyPlus simulated energy can enhance the accuracy of DL models in scenarios where building infrastructure has changed. For instance, if the HVAC system in a building is upgraded, it can significantly impact the energy consumption patterns. However, DL models rely on historical data and cannot inherently account for such specific changes in the building's infrastructure. In contrast, EnergyPlus can be easily updated by modifying the HVAC properties in the software or program to reflect the changes. This generates updated, physically informed forecasts incorporating the effects of the HVAC upgrade. These EnergyPlus predictions can then be used to improve the performance of the deep learning model, effectively bridging the gap between physical changes in the building and the model's predictive capabilities.

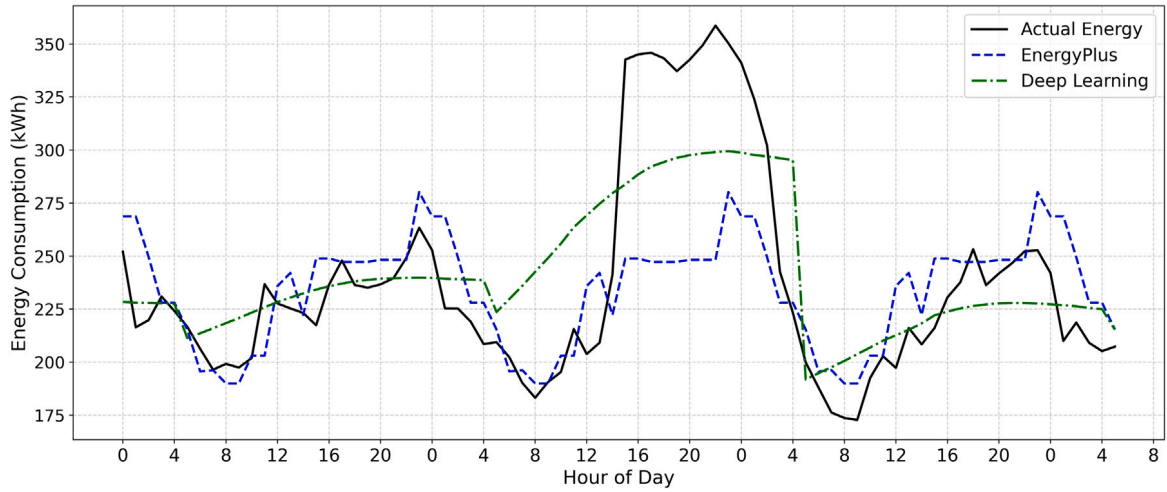

**Fig. 3.** Actual vs. deep learning and EnergyPlus simulated energy consumption (samples from the training period). The deep learning model closely follows the actual energy consumption patterns with smooth variations, whereas EnergyPlus follows the general trend but shows sharper fluctuations and spikes.

**Table 5**
Hyperparameter ranges and selected values.

| Model | Hyperparameter | Range | Selected Value |
|---|---|---|---|
| LSTM | Hidden Dimensions | 32–128 | 64 |
| | Learning Rate | 0.0001–0.01 | 0.001 |
| | Number of Layers | 1–3 | 1 |
| GRU | Hidden Dimensions | 16–64 | 32 |
| | Learning Rate | 0.001–0.1 | 0.01 |
| | Number of Layers | 1–2 | 1 |
| Transformer | Embedding Dimensions | 16–64 | 32 |
| | Number of Heads | 2–8 | 4 |
| | Learning Rate | 0.0001–0.01 | 0.001 |
| | Number of Layers | 1–3 | 1 |

**Table 6**
Performance comparison of models on the test Set.

| Model | MAE | RMSE | SMAPE (%) |
|---|---|---|---|
| LSTM | 23.5 | 26.6 | 10.0 |
| GRU | 23.7 | 26.8 | 10.1 |
| Transformer | 23.5 | 26.6 | 10.0 |

**Table 7**
Performance comparison across five scenarios for the test period using hourly data.

| Scenario | Description | Method | SMAPE (%) | MAE | RMSE |
|---|---|---|---|---|---|
| *Scenario 1:* | Both EnergyPlus and DL available | LSTM | 7.6 | 17.9 | 24.6 |
| | | EnergyPlus | 30.4 | 66.2 | 86.1 |
| | | PgMN | 7.5 | 17.6 | 24.2 |
| *Scenario 2:* | 20% Sparse Ground Truth | LSTM | 46.3 | 62.0 | 111.7 |
| | | EnergyPlus | 30.4 | 66.2 | 86.1 |
| | | PgMN | 44.9 | 59.7 | 107.0 |
| *Scenario 3:* | Actual energy unavailable (new building) | EnergyPlus | 30.4 | 66.2 | 86.1 |
| | | PgMN | 21.2 | 50.2 | 66.0 |
| *Scenario 4:* | No DL predictions, only EnergyPlus | EnergyPlus | 30.4 | 66.2 | 86.1 |
| | | PgMN | 19.3 | 45.0 | 58.0 |
| *Scenario 5:* | No EnergyPlus, only DL predictions | LSTM | 7.6 | 17.9 | 24.6 |
| | | PgMN | 7.5 | 17.8 | 24.3 |

This study explores how the presence of EnergyPlus forecasts and a standalone LSTM can affect the performance of our proposed PgMN approach. In this setup, *we use hourly resolution data spanning an entire year, 8,760 readings in total. From this,* 60% of data was used for EnergyPlus calibration and for DL training, 20% for DL validation, and 20% for testing. We observe how PgMN can improve performance by taking advantage of both. The predictions from both models are given to PgMN as input, and the objective is to improve DL forecasts by taking physically informed knowledge from EnergyPlus prediction. Table 7 and Fig. 4 presents metrics for three comparisons: DL, EP, and our PgMN. The DL method is an LSTM that benefits from historical consumption data, while EP provides a physics-based baseline. PgMN integrates both, attempting to refine the combined forecast further.

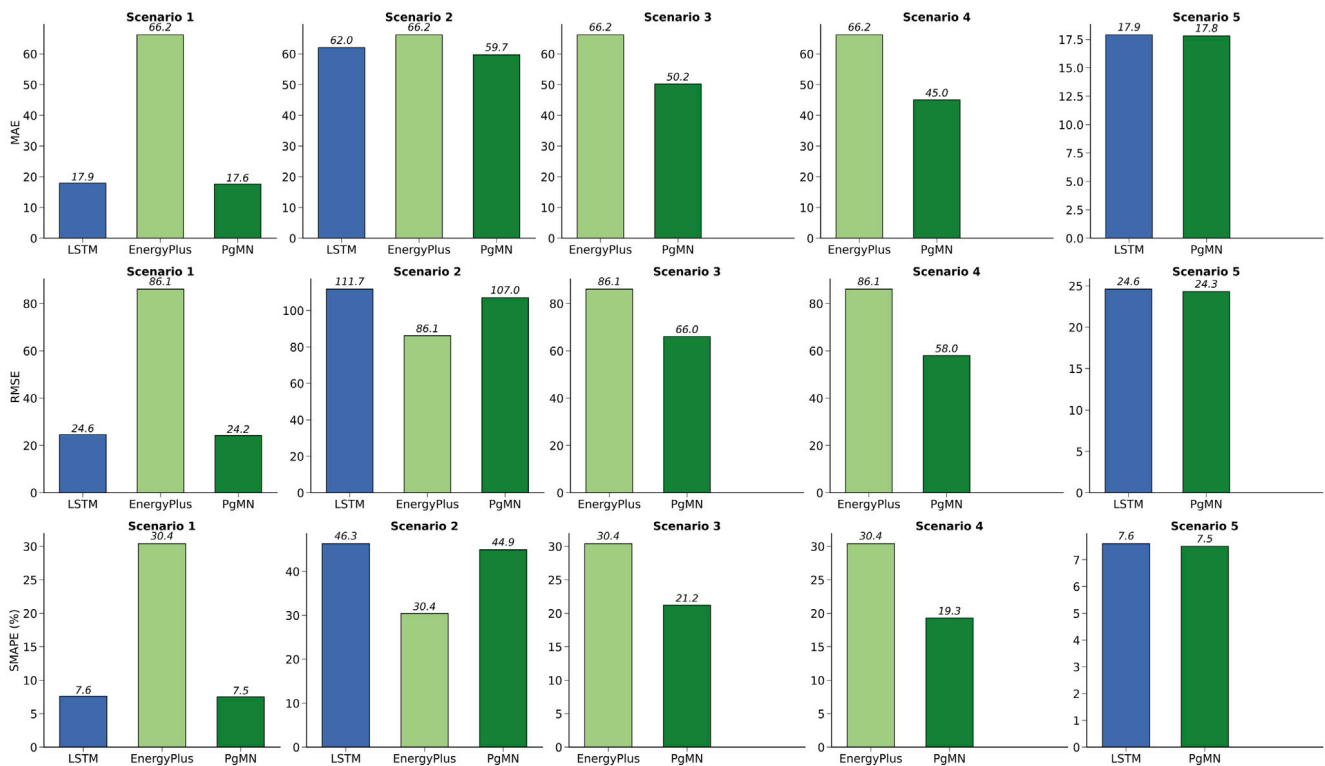

**Fig. 4.** Performance comparison across five scenarios in terms of MAE, RMSE, and SMAPE metrics. Each subplot corresponds to a different scenario, with bars representing LSTM, EnergyPlus, and PgMN results.

As shown in Table 7 and Fig. 4, the standalone LSTM achieves strong results compared to EnergyPlus, yet PgMN slightly outperforms the LSTM in all metrics when both forecasts are used. The SMAPE drops from 7.621% for the LSTM to 7.517% for PgMN, while the RMSE is reduced from 24.591 to 24.209. This indicates that combining physics-based and data-driven outputs can enhance accuracy when both are available.

*5.4.2. Scenario 2: Sparse ground truth data*

In many real-world situations, sensor failures, communication errors, or maintenance downtime often lead to missing data in energy consumption measurements. To simulate such circumstances, 20% of the actual data points are zeroed out and treated as missing. In this scenario, we use both DL and EnergyPlus predictions as inputs for PgMN, dividing the data into 60% for training, 20% for validation, and 20% for testing. The 20% sparsity in actual labels is specifically modified to be treated as missing data, enabling PgMN to learn and adapt under incomplete information conditions.

Our proposed method, PgMN, automatically imputes these missing values using several techniques, including nearest neighbor, linear interpolation, and historical averaging imputation [41], and selects the best approach based on results. The ablation study presented in Table 8 examines performance of PgMN with different imputation techniques. It can be observed that PgMN with linear interpolation performed slightly better than the nearest-neighbor imputation, improving SMAPE from 45.0% to 44.9% and reducing RMSE from 109.7 to 107.0. However, historical averaging led to higher errors, SMAPE of 47.3% and RMSE of 112.1, likely due to its inability to capture short-term local variations. This ablation study confirm that PgMN is flexible regarding the choice of imputation strategy, but the selection of imputation technique further enhances performance in sparse data scenarios.

We compare the performance of LSTM, EnergyPlus, and PgMN under these missing data conditions. Table 7 shows the resulting SMAPE, MAE, and RMSE metrics with linear interpolation when 20% of the actual measurements are unavailable. As observed from the table, the LSTM model shows a drop in performance compared to the sparse actual measurements. The LSTM model utilized three techniques, LI, HA, and NN, to impute missing values and determine which method produced the best results [42]. EnergyPlus remains more robust (SMAPE of 30.412%), relying on its physics-based simulation rather than historical data. PgMN manages a SMAPE of 44.9%, showing that it can still refine the forecast even with limited ground truth. This result stresses the adaptability of our model in handling incomplete sensor readings while benefiting from the physics-driven predictions.

*5.4.3. Scenario 3: Absence of ground truth*

DL-based energy forecasting cannot be applied in newly constructed buildings in many real-world scenarios, as these models require historical consumption data. However, EnergyPlus can be deployed without any prior operational history for the new building. This represents a considerable advantage for EP when ground truth measurements are unavailable (e.g., during the early stages of a building's life cycle). This study evaluates our proposed model without actual measurements (ground truth). Since no data are available, the DL input is replaced with a zero vector (indicating no available DL predictions), and the EP forecasts serve both as the primary input to our model and as the proxy for the unseen actual consumption in the DL training process. Despite this limitation, our method can learn from past EP predictions and improve upon them for the next day (or 24-hour) energy forecasts. Once the model is trained, we show the actual consumption data solely at testing time to verify the model's performance. For this scenario, 60% of EP data is used for training, 20% used for validation, and 20% for testing; the data is prepared using sliding windows, such as the previous 24 h of simulated EP energy used as input, and the model aims to predict the next 24 h of the horizon. Table 7 summarizes the comparative results between EnergyPlus and PgMN.

**Table 8**
Comparison of imputation strategies used in PgMN for handling sparse ground truth (20% missing data).

| Method | SMAPE (%) | MAE | RMSE |
| --- | --- | --- | --- |
| PgMN with Nearest Neighbor | 45.0 | 60.4 | 109.7 |
| PgMN with Historical Averaging | 47.3 | 62.9 | 112.1 |
| PgMN with Linear Interpolation | **44.9** | **59.7** | **107.0** |

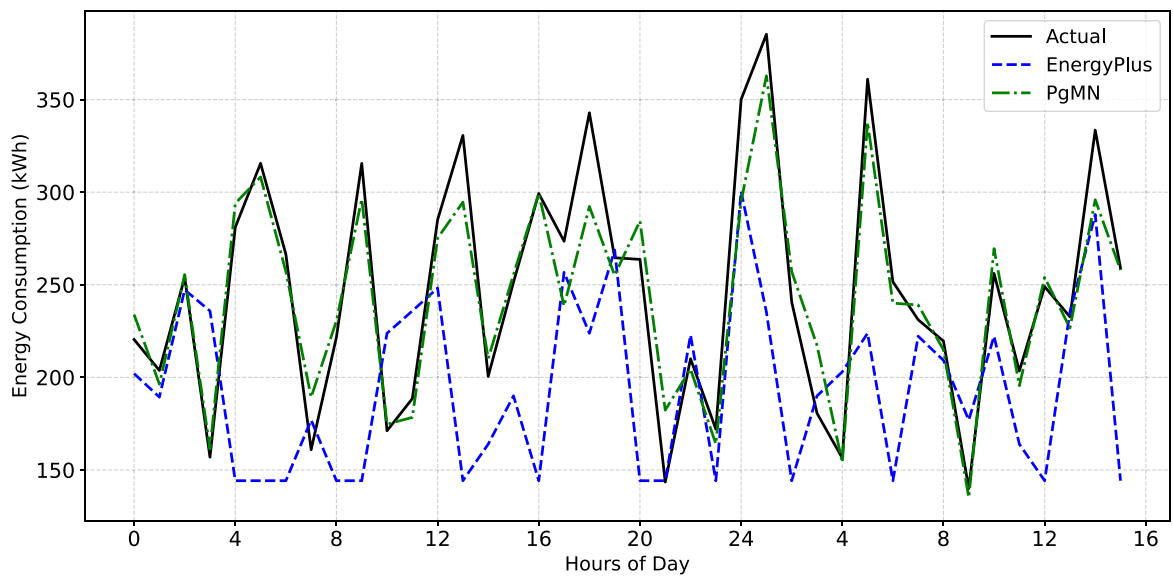

**Fig. 5.** Scenario 3 – Comparison of actual vs. energy predicted by PgMN. In the absence of ground truth during training, EnergyPlus simulated energy is guiding PgMN to follow the physically informed pattern to improve accuracy, such as during the second day or 30 to 40 h.

As seen in Table 7, EnergyPlus alone results in SMAPE of 30.412%, MAE of 66.236, and RMSE of 86.121. Our proposed approach PgMN, even when forced to use zeros in place of DL forecasts and relying on EP predictions as the training target, shows marked improvement with a SMAPE of 21.22%, MAE of 50.212, and RMSE of 66.012. This stresses the possibility of leveraging physics-based simulations as a substitute for actual data in early building operation phases while still allowing the model to refine the raw EP predictions, but zeroing DL is really important considering our model design, which allows for learning from wide margin such as from zeros to EnergyPlus predictions which significantly reduces overall error by almost 8%, and this can be highly applicable situation for this case. Fig. 5 shows actual vs predicted energy consumption values, whereas EnergyPlus predictions are also given to understand how well the proposed method is able to learn complex patterns.

*5.4.4. Scenario 4: Sole utilization of physics-based EnergyPlus*

In many real-world settings, only EnergyPlus forecasts may be available to predict energy consumption. This commonly occurs when DL forecasts cannot be produced due to a lack of historical data, often in newly constructed or unique buildings. In this study, we replicate this scenario by setting the DL input to zeros, allowing the model to operate solely on EnergyPlus outputs. Despite having no DL signals, the model refines the raw EP predictions by learning patterns from past behavior. In this situation, we allocate 60% of the EnergyPlus data for training, 20% for validation, and 20% for testing. While DL inputs are provided, they contain zero values, which allows PgMN to explore a broader range to find the optimal point between the predictions from DL and EnergyPlus, where the loss is minimized.

As seen in Table 7 and Fig. 6, EnergyPlus alone achieves a SMAPE of 30.412%, MAE of 66.236, and RMSE of 86.121. PgMN reduces these values to a SMAPE of 19.390%, MAE of 45.099, and RMSE of 58.039, stressing its capacity to enhance forecasts by leveraging physics-based data without any deep learning predictions.

*5.4.5. Scenario 5: Sole utilization of deep learning*

Some buildings may not be suitable for EnergyPlus models due to the unavailability of building parameters, or the simulation outputs may be deemed insufficient for forecasting. In these circumstances, DL remains the primary tool for data-driven predictions. The PgMN will still function by zeroing out EnergyPlus predictions and relying solely on DL predictions. The DL forecasts are split into 60% for training, 20% for validation, and 20% for testing. While EnergyPlus predictions are also provided as input, they are set to zero. Similar to Scenario 4, PgMN will identify the optimal balance between the two inputs where the error is minimized. In this case, actual energy labels are provided. The goal is to assess PgMN's capacity to refine existing deep learning outputs in the absence of other simulation data.

As seen in Table 7, the original DL model already delivers relatively low error rates. Nevertheless, PgMN consistently improves these metrics by leveraging its internal memory mechanism and the transformation of input signals, achieving a slight but meaningful reduction in SMAPE from 7.621% to 7.567%, and a decrease in RMSE from 24.591 to 24.302. This stresses PgMN's ability to enhance existing deep learning forecasts, even without any additional physics-based information.

*5.5. Ablation study*

To validate the contribution of the Memory Unit, we conducted an ablation study under Scenario 1, where both EnergyPlus and DL predictions were available. The Memory Unit is a learnable vector that records historical forecasting deviations in a trainable manner, capturing persistent biases that arise between physics-based simulated energy and DL predictions. Table 9 presents the results of the ablation study comparing pgMN with and without Memory Unit. When Memory Unit is removed, PgMN performance degrades, with the SMAPE increasing from 7.5% to 7.7% and RMSE from 24.2 to 25.4. This emphasizes that the Memory Unit enables PgMN to dynamically adjust predictions

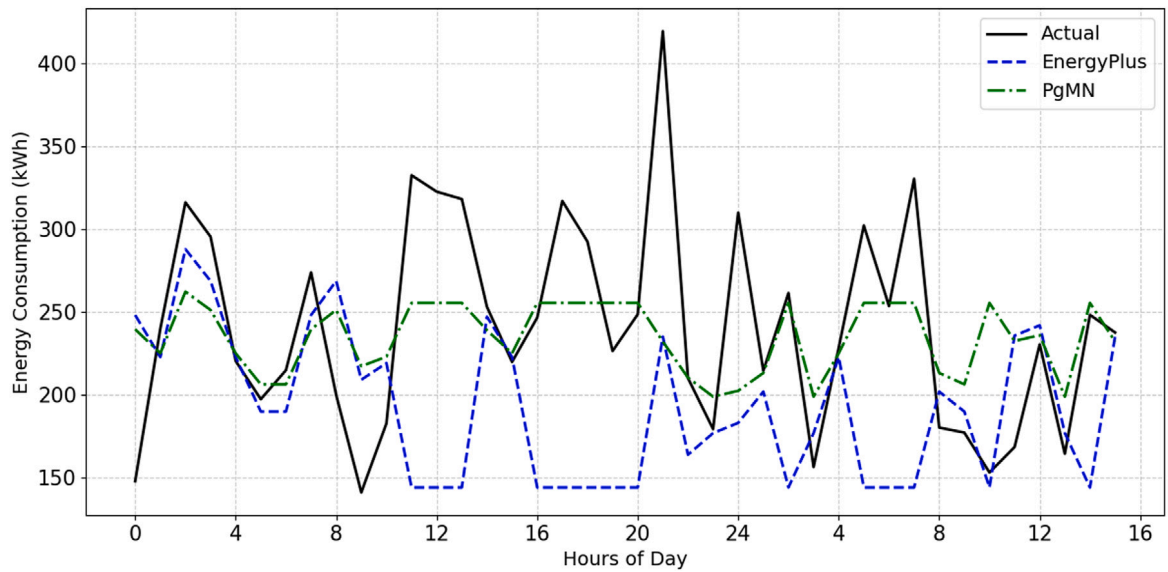

**Fig. 6.** Scenario 4– Comparison of actual vs. energy predicted by PgMN using only EnergyPlus inputs. The sole utilization of simulated energy is helping PgMN to follow patterns, as evidenced between the 0th and 10th hours of the day.

**Table 9**
Ablation study evaluating the contribution of the Memory Unit (Scenario 1).

| Method | SMAPE (%) | MAE | RMSE |
| --- | --- | --- | --- |
| LSTM | 7.6 | 17.9 | 24.6 |
| EnergyPlus | 30.4 | 66.2 | 86.1 |
| **PgMN with Memory Unit** | **7.5** | **17.6** | **24.2** |
| PgMN without Memory Unit | 7.7 | 18.1 | 25.4 |

by learning from historical bias patterns rather than relying solely on feature representations.

To further bridge the theoretical and practical connection of Theorem 4.2, we conduct a bias reduction analysis under Scenario 1. We evaluate the *Mean Error*, defined as the average of the signed differences between the true energy consumption and the predicted values:

Mean Error = $\frac{1}{N} \sum_{i=1}^{N} (y_i - \hat{y}_i)$ where $y_i$ and $\hat{y}_i$ denote the true and predicted energy consumption, respectively. Table 10 presents a sample-wise comparison: the Memory Unit consistently reduces signed errors (e.g., from −17.21 kWh to −5.63 kWh in the first sample), taking forecasts closer to zero bias and mitigating both under- and over-prediction. This detailed analysis validates that the Memory Unit effectively learns and corrects global forecasting deviations, enhancing the model's ability in practice.

### 5.6. Discussion

This subsection discusses the computational time, system-level bias handling, deployment challenges, and practical scalability of PgMN, comprehending its strengths and limitations for real-world applications. To evaluate the computational efficiency of the proposed PgMN, we measured the training and testing times across various scenarios. The training time represents the total time to train the model using early stopping, while the testing time represents the time required to evaluate the model on the test dataset. We conducted experiments on a workstation equipped with an AMD Ryzen Threadripper PRO 5955WX processor and an NVIDIA GA102GL RTX A6000 GPU. The training time remained consistent across all scenarios, averaging approximately 3.5 s, while the testing time was around 0.02 s. The testing time is minimal across all scenarios, emphasizing the PgMN efficiency during inference. This shows that PgMN is computationally efficient and can be deployed in real-world situations.

Nevertheless, it is important to discuss the modeling steps required to setup the PgMN model before it is ready for inference, including obtaining forecasts from a data-driven model and simulated energy from a physics-based model. Setting up EnergyPlus requires significant time and expertise. We first modeled the building in EnergyPlus and calibrated it strictly following ASHRAE Guideline 14-2014. Designing the complete building model, including specifying material properties, glazing details, HVAC system specifications, and schedules, took approximately 10 h. The subsequent calibration process that involves iterative adjustment of simulation inputs to meet ASHRAE accuracy thresholds required an additional 22 h. After calibration, a complete simulation run of the building model took approximately 2.5 h. Following the EnergyPlus modeling, we trained our selected LSTM model on two years of historical hourly data, which required 39.9 min for training and 0.13 s for full-test inference. It is important to note that although EnergyPlus modeling demands significant setup time, it remains the standard practice for accurate energy consumption modeling, particularly for new construction projects. In contrast, the LSTM model training and testing are relatively fast. In total, approximately 35 h were required for the EnergyPlus modeling and DL training phases before PgMN could begin its computations.

It is important to acknowledge that the considerable modeling time, expert knowledge required for EnergyPlus configuration, and dependence on detailed building parameters can limit its feasibility in large-scale urban deployments. Moreover, EnergyPlus simulations are computationally intensive, usually requiring high modeling times that can present latency constraints, making them impractical for real-time forecasting across numerous buildings. These computational trade-offs, including modeling complexity and simulation delays, must be considered when scaling to broader urban settings. However, if historical data are available, PgMN can rely on the DL component, without using EnergyPlus. Nevertheless, for newly constructed buildings where historical data are unavailable, EnergyPlus remains the only practical solution, as DL models depend on historical data for training and cannot operate otherwise. PgMN is designed for both conditions: when EnergyPlus predictions are available, it leverages them to enhance forecasting accuracy; when EnergyPlus is infeasible, PgMN maintains functionality by relying solely on DL forecasts. Moreover, the architecture flexibly handles partial and missing inputs, ensuring operational robustness under practical constraints. Future work will explore reduced-order physics models or precomputed EnergyPlus simulations to minimize computational demands and latency for scalable, real-world deployment.

**Table 10**
Comparison of prediction errors with and without the Memory Unit under Scenario 1. Lower absolute signed error is highlighted in bold (only least error inside parentheses is bold).

| DL | EP | Actual Energy | PgMN (With MU) | PgMN (Without MU) |
| --- | --- | --- | --- | --- |
| 242.29 | 256.71 | 253.31 | 247.68 (−**5.63**) | 236.10 (−17.21) |
| 283.73 | 223.84 | 297.02 | 287.10 (−9.92) | 288.67 (−**8.35**) |
| 266.39 | 144.21 | 262.12 | 250.91 (−**11.21**) | 228.58 (−33.54) |
| 390.44 | 335.24 | 343.41 | 353.42 (+**10.01**) | 355.99 (+12.58) |
| 222.80 | 223.84 | 237.77 | 233.34 (−4.43) | 234.58 (−**3.19**) |
| 252.08 | 144.21 | 237.85 | 148.80 (−**89.05**) | 146.01 (−91.84) |
| 139.40 | 144.21 | 150.64 | 146.70 (−**3.94**) | 145.79 (−4.85) |
| 222.64 | 203.01 | 187.41 | 191.79 (+**4.38**) | 193.02 (+5.61) |
| 163.00 | 195.55 | 163.28 | 176.01 (+**12.73**) | 177.65 (+14.37) |
| 253.21 | 144.21 | 251.54 | 148.80 (−102.74) | 150.05 (−**101.49**) |
| 146.00 | 209.20 | 141.21 | 198.14 (+**56.93**) | 199.41 (+58.20) |
| 216.00 | 144.21 | 230.36 | 236.12 (+**5.76**) | 240.60 (+10.24) |
| 219.51 | 235.24 | 224.12 | 234.08 (+**9.96**) | 235.49 (+11.37) |
| 182.79 | 144.21 | 129.52 | 143.29 (+**13.77**) | 144.27 (+14.75) |
| 164.00 | 202.02 | 162.04 | 194.23 (+32.19) | 193.07 (+**31.03**) |
| 168.00 | 202.02 | 169.12 | 194.89 (+**25.77**) | 196.45 (+27.33) |
| … | … | … | … | … |
| | | Mean Error | **17.6** | 18.1 |

While DL and physics-based streams in Parallel Project Layer (Fig. 1) provide complementary information $x_d$ and $x_e$, both are generated independently and inherently contain prediction errors due to data limitations (for DL) and modeling assumptions (for EnergyPlus). Instead of correcting errors individually, PgMN integrates them at the system level. The Memory Unit captures persistent bias patterns across forecasts, while the Memory Experience Module dynamically learns optimal sample-specific combinations of $x_d$ and $x_e$, or extrapolations beyond their range, as necessary to minimize forecast error. This architecture reduces cumulative prediction errors and improves model adaptability across different operational scenarios.

This study represents a real-world, industry-aided project, where both historical energy consumption data and access to detailed building design for EnergyPlus modeling were available. This setup allowed us to compare different realistic scenarios. However, this dual data availability is challenging and presents a major challenge. In most real-world cases, detailed building models or extensive historical measurements are unavailable, limiting the scenarios that could be considered in PgMN. Furthermore, this project focused on a large residence building where occupant patterns are somewhat regular, a factor difficult to record for other buildings. To further examine PgMN generalizability, future work will evaluate PgMN across diverse building types, climates, and consumer usage patterns.

## 6. Conclusion

This paper presents the PgMN, a neural network that integrates the strengths of DL and PBMs to enhance energy consumption forecasting. PgMN addresses limitations of standalone DL and PBMs, such as DL's dependency on historical data and PBM's rigidity, such as vast requirements of building parameters. The PgMN employs a theoretically proven Parallel Projection Layer, Memory Unit, and Memory Experience Module and can handle incomplete data, dynamically correct forecasting biases, and optimally combine or extend predictions. We conducted extensive experiments across diverse scenarios to validate PgMN's robustness and accuracy. PgMN uses PBM outputs to produce reliable forecasts when actual data is sparse. Similarly, without PBM predictions, PgMN refines DL outputs through its memory and aggregation mechanisms. The proposed model improves forecasting accuracy and reduces dependency on a single predictive approach, making it highly applicable to newly constructed buildings, infrastructure changes, and dynamic consumer behavior.

Future research will explore extending PgMN to other domains requiring hybrid models, such as HVAC optimization and intelligent grid management, further expanding PgMN's versatility and impact.

## CRediT authorship contribution statement

**Muhammad Umair Danish:** Writing – review & editing, Writing – original draft, Visualization, Validation, Software, Resources, Methodology, Investigation, Formal analysis, Data curation. **Kashif Ali:** Methodology, Data curation. **Kamran Siddiqui:** Writing – review & editing, Methodology, Funding acquisition, Conceptualization. **Katarina Grolinger:** Writing – review & editing, Validation, Supervision, Resources, Project administration, Investigation, Funding acquisition, Data curation, Conceptualization.

## Declaration of competing interest

The authors declare the following financial interests/personal relationships which may be considered as potential competing interests: Katarina Grolinger reports financial support was provided by work was supported in part by the Climate Action and Awareness Fund [EDF-CA-2021i018, Environnement Canada, K.Siddiqui and K. Grolinger] and in part by the Canada Research Chairs Program [CRC-2022-00078, K. Grolinger]. If there are other authors, they declare that they have no known competing financial interests or personal relationships that could have appeared to influence the work reported in this paper.

## Data availability

The data that has been used is confidential.

## References

[1] Guo Y, Li Y, Qiao X, Zhang Z, Zhou W, Mei Y, et al. Bilstm multitask learning-based combined load forecasting considering the loads coupling relationship for multienergy system. IEEE Trans Smart Grid 2022.

[2] ZipDo Research. AI in the electrical industry: Statistics and insights. 2021, URL https://zipdo.co/ai-in-the-electrical-industry-statistics/. [Accessed 25 November 2024].

[3] Saxena A, Shankar R, El-Saadany E, Kumar M, Al Zaabi O, Al Hosani K, et al. Intelligent load forecasting and renewable energy integration for enhanced grid reliability. IEEE Trans Ind Appl 2024.

[4] Real AC, Luz GP, Sousa J, Brito M, Vieira S. Optimization of a photovoltaic-battery system using deep reinforcement learning and load forecasting. Energy AI 2024;16:100347.

[5] Nti IK, Teimeh M, Nyarko-Boateng O, Adekoya AF. Electricity load forecasting: a systematic review. J Electr Syst Inf Technol 2020.

[6] Singh SK. Hybrid machine learning approach for predictive modeling of complex systems [Ph.D. thesis], State University of New York at Buffalo; 2019.

[7] Jia X, Karpatne A, Willard Jea. Physics guided recurrent neural networks for modeling dynamical systems: Application to monitoring water temperature and quality in lakes. 2018, arXiv preprint arXiv:1810.02880.

[8] Tobin J, Fong R, Ray A, et al. S. Domain randomization for transferring deep neural networks from simulation to the real world. In: International conference on intelligent robots and systems. IEEE; 2017.

[9] Lv L, Wu Z, Zhang L, Gupta BB, Tian Z. An edge-AI based forecasting approach for improving smart microgrid efficiency. IEEE Trans Ind Inform 2022.

[10] Zhang R, Mirzaei PA. Virtual dynamic coupling of computational fluid dynamics-building energy simulation-artificial intelligence: Case study of urban neighbourhood effect on buildings' energy demand. Build Environ 2021.

[11] Ma Z, Jiang G, Hu Y, Chen J. A review of physics-informed machine learning for building energy modeling. Appl Energy 2025.

[12] Zerrougui I, Li Z, Hissel D. Physics-informed neural network for modeling and predicting temperature fluctuations in proton exchange membrane electrolysis. Energy AI 2025.

[13] Lee H, Lee C, Lee H. Physics-informed machine learning for enhanced prediction of condensation heat transfer. Energy AI 2025;100482.

[14] Magri L, Doan NAK. First-principles machine learning modelling of COVID-19. 2020, arXiv preprint arXiv:2004.09478.

[15] Kamrava S, Tahmasebi P, Sahimi M. Enhancing images of shale formations by a hybrid stochastic and deep learning algorithm. Neural Netw 2019.

[16] Pizzati F, Cerri P, de Charette R. Physics-informed guided disentanglement in generative networks. IEEE Trans Pattern Anal Mach Intell 2023.

[17] Meng Y, Rigall E, Chen Xea. Physics-guided generative adversarial networks for sea subsurface temperature prediction. IEEE Trans Neural Netw. Learn Syst 2021.

[18] Chen Y, Yang Q, Chen Z, Yan C, Zeng S, Dai M. Physics-informed neural networks for building thermal modeling and demand response control. Build Environ 2023.

[19] Bento ME. Physics guided neural network for load margin assessment of power systems. IEEE Trans Power Syst 2023.

[20] Hofmann T, Hamar J, Mager B, Erhard S, Schmidt JP. Physics-constrained transfer learning: Open-circuit voltage curve reconstruction and degradation mode estimation of lithium-ion batteries. Energy AI 2025;100493.

[21] Nutkiewicz A, Yang Z, Jain RK. Data-driven urban energy simulation (DUE-s): A framework for integrating engineering simulation and machine learning methods in a multi-scale urban energy modeling workflow. Appl Energy 2018.

[22] Xiao T, You F. Building thermal modeling and model predictive control with physically consistent deep learning for decarbonization and energy optimization. Appl Energy 2023.

[23] Tian C, Ye Y, Lou Y, Zuo W, Zhang G, Li C. Daily power demand prediction for buildings at a large scale using a hybrid of physics-based model and generative adversarial network. In: Building simulation. Springer; 2022.

[24] Ma Z, Jiang G, Chen J. Physics-informed ensemble learning with residual modeling for enhanced building energy prediction. Energy Build 2024.

[25] Brøgger M, Bacher P, Wittchen KB. A hybrid modelling method for improving estimates of the average energy-saving potential of a building stock. Energy Build 2019.

[26] Jiang Z, Dong B. Modularized neural network incorporating physical priors for future building energy modeling. Patterns 2024.

[27] Quick Start Energy Plus Guide. https://energyplus.readthedocs.io/en/latest/quick_start/quick_start.html.[Accessed 25 January 2024].

[28] Zhu D, Hong T, Yan D, Wang C. A detailed loads comparison of three building energy modeling programs: EnergyPlus, DeST and DOE-2.1 e. In: Building simulation. Springer; 2013.

[29] Quick Start Open Studio Guide. 2024, https://openstudio.net/.[Accessed 25 January 2024].

[30] Design Builder Energy Modeling Tool. 2024, https://designbuilder.co.uk/.[Accessed 25 January 2024].

[31] Seyyedi A, Bohlouli M, Oskoee SN. Machine learning and physics: A survey of integrated models. ACM Comput Surv 2023.

[32] Danish MU, Grolinger K. Leveraging hypernetworks and learnable kernels for consumer energy forecasting across diverse consumer types. IEEE Trans Power Deliv 2024.

[33] Wang G, Giannakis GB, Chen J. Learning ReLU networks on linearly separable data: Algorithm, optimality, and generalization. IEEE Trans Signal Process 2019.

[34] Evans R, Bošnjak M, Buesing L, Ellis K, Pfau D, Kohli P, et al. Making sense of raw input. Artificial Intelligence 2021;299:103521.

[35] Lu L, Jin P, Pang G, Zhang Z, Karniadakis GE. Learning nonlinear operators via DeepONet based on the universal approximation theorem of operators. Nat Mach Intell 2021.

[36] Simonyan K, Zisserman A. Very deep convolutional networks for large-scale image recognition. In: 3rd international conference on learning representations (ICLR 2015). Computational and Biological Learning Society; 2015.

[37] Ashrae. ASHRAE guideline 14-2014: Measurement of energy, demand, and water savings. 2014, https://webstore.ansi.org/standards/ASHRAE/ashraeguideline142014. [Accessed 22 April 2025].

[38] Government of Ontario. Ontario's building code. 2024, URL https://www.ontario.ca/page/ontarios-building-code. [Accessed 22 April 2025].

[39] Lee K, Lim H. Correlation analysis of building parameters according to ashrae standard 90.1. J Build Eng 2024;82:108130.

[40] Deng L. A tutorial survey of architectures, algorithms, and applications for deep learning. APSIPA Trans Signal Inf Process 2014.

[41] Ryu S, Kim M, Kim H. Denoising autoencoder-based missing value imputation for smart meters. IEEE Access 2020;8:40656–66.

[42] Li L, Zhang J, Wang Y, Ran B. Missing value imputation for traffic-related time series data based on a multi-view learning method. IEEE Trans Intell Transp Syst 2018.